\documentclass[sigconf]{acmart}%, review
\usepackage{multirow}
\usepackage{booktabs}
\usepackage{amsfonts} 
\usepackage{bm}
\usepackage{CJKutf8}
\usepackage{caption}
\usepackage{subfigure}
\usepackage{booktabs}
\usepackage{float}

\newcommand{\ie}{\textit{i.e.}}
\newcommand{\eg}{\textit{e.g.}}

\copyrightyear{2021} 
\acmYear{2021} 
\setcopyright{acmlicensed}\acmConference[SIGIR '21]{Proceedings of the 44th International ACM SIGIR Conference on Research and Development in Information Retrieval}{July 11--15, 2021}{Virtual Event, Canada}
\acmBooktitle{Proceedings of the 44th International ACM SIGIR Conference on Research and Development in Information Retrieval (SIGIR '21), July 11--15, 2021, Virtual Event, Canada}
\acmPrice{15.00}
\acmDOI{10.1145/3404835.3463239}
\acmISBN{978-1-4503-8037-9/21/07}

\acmSubmissionID{1212}

\begin{document}
\fancyhead{}

%%
%% The "title" command has an optional parameter,
%% allowing the author to define a "short title" to be used in page headers.
\title{Pchatbot: A Large-Scale Dataset for Personalized Chatbot}

%%
%% The "author" command and its associated commands are used to define
%% the authors and their affiliations.
%% Of note is the shared affiliation of the first two authors, and the
%% "authornote" and "authornotemark" commands
%% used to denote shared contribution to the research.
% \author{Hongjin Qian}
% \email{ian@ruc.edu.cn}
% \orcid{1234-5678-9012}
% \affiliation{%
%   \institution{Renmin University of China}
%   \city{Beijing}
% }

\author{Hongjin Qian\textsuperscript{\rm 1}, Xiaohe Li\textsuperscript{\rm 2}, Hanxun Zhong\textsuperscript{\rm 2}, Yu Guo\textsuperscript{\rm 2}, Yueyuan Ma\textsuperscript{\rm 2}, Yutao Zhu\textsuperscript{\rm 3}\\
Zhanliang Liu\textsuperscript{\rm 2}, Zhicheng Dou\textsuperscript{\rm 1*}, Ji-Rong Wen\textsuperscript{\rm 1,2}}
\affiliation{
\textsuperscript{\rm 1} Gaoling School of Artificial Intelligence, Renmin University of China \state{Beijing} \country{China} \\
\textsuperscript{\rm 2} School of Information, Renmin University of China \state{Beijing} \country{China} \\
\textsuperscript{\rm 3} Université de Montréal \state{Québec} \country{Canada}}
%\email{
%{ian, lixiaohe, hanxun\_zhong, mayueyuan2016, yu\_guo}@ruc.edu.cn}
%\email{
%yutao.zhu@umontreal.ca, 
%{zliu, dou, jrwen}@ruc.edu.cn 
%}
\email{{ian, *dou}@ruc.edu.cn}
\renewcommand{\authors}{Hongjin Qian, Xiaohe Li, Hanxun Zhong, Yu Guo, Yueyuan Ma, Yutao Zhu, Zhanliang Liu, Zhicheng Dou, and Ji-Rong Wen}

%%
%% By default, the full list of authors will be used in the page
%% headers. Often, this list is too long, and will overlap
%% other information printed in the page headers. This command allows
%% the author to define a more concise list
%% of authors' names for this purpose.
\renewcommand{\shortauthors}{Qian, et al.}

%%
%% The abstract is a short summary of the work to be presented in the
%% article.
\begin{abstract}
Natural language dialogue systems raise great attention recently. As many dialogue models are data-driven, high-quality datasets are essential to these systems. In this paper, we introduce Pchatbot, a large-scale dialogue dataset that contains two subsets collected from Weibo and Judicial forums respectively. To adapt the raw dataset to dialogue systems, we elaborately normalize the raw dataset via processes such as anonymization, deduplication, segmentation, and filtering. The scale of Pchatbot is significantly larger than existing Chinese datasets, which might benefit the data-driven models. Besides, current dialogue datasets for personalized chatbot usually contain several persona sentences or attributes. Different from existing datasets, Pchatbot provides anonymized user IDs and timestamps for both posts and responses. This enables the development of personalized dialogue models that directly learn implicit user personality from the user's dialogue history. Our preliminary experimental study benchmarks several state-of-the-art dialogue models to provide a comparison for future work. The dataset can be publicly accessed at Github: \url{https://github.com/qhjqhj00/Pchatbot}.

%Our preliminary experimental study shows that a personalized chatbot model trained on Pchatbot outperforms the corresponding ad-hoc chatbot models\dou{this is not our goal. Our target is to prove the quality of the dataset}. We also demonstrate that using larger dataset improves the quality of dialog models. 

%\dou{you need to mention which kind of efforts you made on processing this dataset; you also need to mention its value}
\end{abstract}


%%
%% The code below is generated by the tool at http://dl.acm.org/ccs.cfm.
%% Please copy and paste the code instead of the example below.
%%
%  \begin{CCSXML}
% <ccs2012>
%  <concept>
%   <concept_id>10010520.10010553.10010562</concept_id>
%   <concept_desc>Computer systems organization~Embedded systems</concept_desc>
%   <concept_significance>500</concept_significance>
%  </concept>
%  <concept>
%   <concept_id>10010520.10010575.10010755</concept_id>
%   <concept_desc>Computer systems organization~Redundancy</concept_desc>
%   <concept_significance>300</concept_significance>
%  </concept>
%  <concept>
%   <concept_id>10010520.10010553.10010554</concept_id>
%   <concept_desc>Computer systems organization~Robotics</concept_desc>
%   <concept_significance>100</concept_significance>
%  </concept>
%  <concept>
%   <concept_id>10003033.10003083.10003095</concept_id>
%   <concept_desc>Networks~Network reliability</concept_desc>
%   <concept_significance>100</concept_significance>
%   </concept>
%  </ccs2012>
%  \end{CCSXML}

% \ccsdesc[500]{Computer systems organization~Embedded systems}
% \ccsdesc[300]{Computer systems organization~Redundancy}
% \ccsdesc{Computer systems organization~Robotics}
% \ccsdesc[100]{Networks~Network reliability}

%%
%% Keywords. The author(s) should pick words that accurately describe
%% the work being presented. Separate the keywords with commas.

%% A "teaser" image appears between the author and affiliation
%% information and the body of the document, and typically spans the
%% page.

%%
%% This command processes the author and affiliation and title
%% information and builds the first part of the formatted document.
\maketitle

\section{Introduction}
The dialogue system is a longstanding challenge in Artificial Intelligence. Intelligent dialogue agents have been rapidly developed but their effectiveness is still far behind general expectations. The reasons for the lag are multi-dimensional in which the lack of datasets is a fundamental constraint. Training a chatbot usually requires a large-scale dataset, but collecting real conversations between people requires tremendous human labor. Hence, most existing studies mainly leverage publicly available post-comments to simulate conversations between users. Example datasets are Ubuntu Dialogue Corpus~\cite{Lowe_UbuntuV2_2015} and Douban Corpus~\cite{Wu_Douban_2017}, which are sourced from online forums. As discussed by \citet{gao2019neural}, there are still many challenges for dialogue chatbot. Personality consistency is one of these challenges. Regarding personalized dialogue dataset, previous works depict personality using either personality sentences (PERSONA-CHAT~\cite{Zhang_dog_2018}) or persona attributes (Personality Assignment Dataset~\cite{Huang_Chat_2018}). With the availability of these datasets, lots of dialogue models have been proposed. 

% \dou{you need to mention their sizes}

% \dou{there is a sematic gap between previous sentences and the next sentence (personalized dialogue dataset is a new concept). You need to mention that to solve ... problem, there are more efforts on developing personalized chatbot which try to maintain a consistent personality for a chatbot. ....}

Many recent neural dialogue models demonstrate substantial gains on chatbots' performances using large-scale dataset~\cite{Samuel_fb_2018,zhang2020dialogpt,brown2020language}. Thus, the scale of most current dialogue datasets becomes a major barrier that limits the development of neural dialogue models. The limitation of the data scale is exacerbated regarding the personalized dialogue dataset. The main reasons include: (1) Most current personalized datasets are based on the explicit user profiles. Such profiles are often manually annotated (\eg, user's persona descriptions), thus being very costly. 
(2) Explicit user profiles can only provide relatively limited personalized information as explicit user profiles are usually comprised of static personality descriptions. For example, in PERSONA-CHAT~\cite{Zhang_dog_2018}, each interlocutor is described by five short sentences. Such static property also makes the explicit user profiles hard to update.

To tackle the aforementioned problems, in this work, we introduce Pchatbot, a large-scale Chinese conversation dataset dedicated to the development of personalized dialogue models. Pchatbot has two subsets, named PchatbotW and PchatbotL, built from open-domain Weibo and several judicial forums. They respectively have 130 million and 59 million high-quality conversations. To the best of our knowledge, Pchatbot is the largest Chinese dialogue dataset. As shown in Table~\ref{tab:scale}, PchatbotW is 14 times larger than the current largest persona-based open-domain corpus in Chinese, \ie, Personality Assignment Dataset~\cite{Huang_Chat_2018}. The detailed statistics of Pchatbot are shown in Table~\ref{tab:stat_Pchatbot}.

% \dou{I don't want this. I can be used for general chatbots, and you need to propose some other tasks}.

\begin{figure}[t!]
    \centering
    \includegraphics[width=0.45\textwidth]{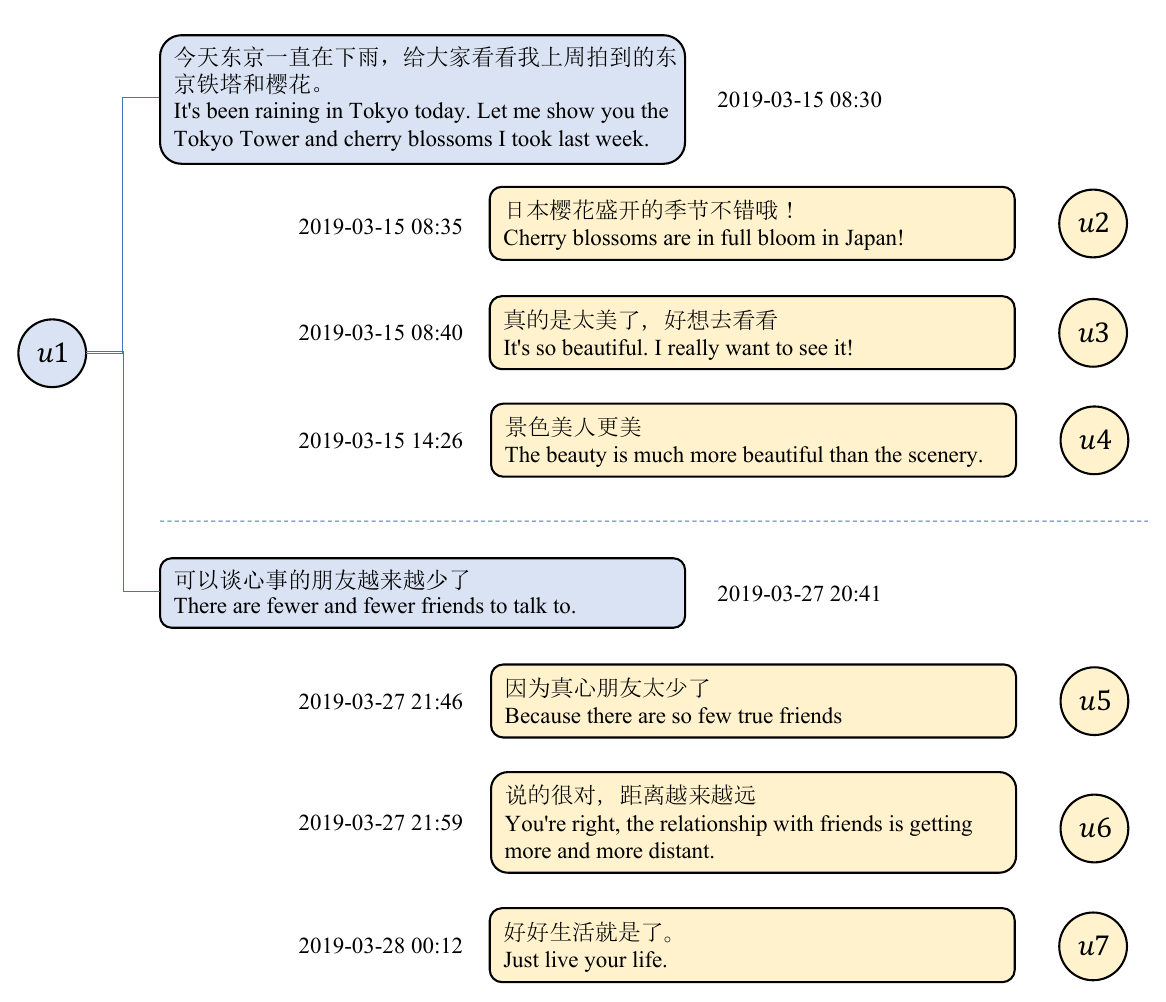}
    \caption{Examples of PchatbotW.}
    \label{fig:sample1}
    \vspace{-10px}
\end{figure}

An example of PchatbotW is illustrated in Figure~\ref{fig:sample1}. In addition to the content of the post-response pairs, the ID of the corresponding user and the publication timestamp are also provided. We believe such data have the potential to support several kinds of research questions, at least the following three: (1) \textit{Single-turn dialogue}. This task only considers the interaction between the post and the response within a single turn~\cite{Shang_Noah_2015,Personal_turning_VinyalsL15,Jiwei_distinct_2016,DBLP:conf/aaai/XingWWLHZM17,DBLP:journals/ir/ZhuDNW20}, thus being naturally supported by our dataset. As our dataset is significantly larger than other Chinese datasets, it is promising to learn a model with better performance~\cite{Samuel_fb_2018,zhang2020dialogpt,brown2020language}. (2) \textit{Multi-referenced dialogue}. In natural language dialogue, a post can have multiple appropriate responses. So, recent studies start to explore generating diverse responses for a given input~\cite{DBLP:conf/acl/QiuLBZY19,DBLP:journals/corr/abs-2009-07117,DBLP:conf/emnlp/LachauxJL20}. It is evident to see that our dataset can also support this kind of research since several responses corresponding to one post are collected. (3) \textit{Personalized dialogue}. With the help of user IDs and timestamps, we can aggregate user-wise data to obtain the user's dialogue history. As tremendous personalized information (such as speaking style, vocabulary, and interests) is hidden under the user's dialogue history, intuitively we can design models that directly learn implicit user profiles from the dialogue history, which enlightens a new research way for personalized chatbots. Besides, the Pchatbot dataset has user ID for both sides of interlocutors, which expands the application scenarios where we can model the personalities of both the source user and the target user. In this work, we mainly focus on investigating the application of our dataset on personalized dialogue. 
% With such bilateral personalized information, dialogue models can be further designed for one-to-many personalized dialogue.
% To protect user privacy, data anonymization processes are applied. 

%\dou{about privacy, you need to check previous review comments. Reviewers argue that removing these information doesn't mean you deal with all privacy problem. You need to mention that all data we collected are publicly available on the original websites, and we don't include any private conversations between users. So even if we have users Ids, we don't expose more private information of users}

Dataset construction is usually confronted with two major challenges: (1) how to adapt the raw data to a specific task, and (2) how to protect user privacy. The most intuitive idea addressing the two challenges is to directly remove data noise or private information. However, we need to consider the scope of data noise or private information and how to recognize them. Furthermore, whether removing such texts undermines semantics also remains unsolved. After fine-grained data analysis and human evaluation, we conclude that the quality of the dataset can be effectively improved via removing meaningless conversations including hashtags, URLs, emoticons, text duplication, and multi-languages. Besides, yet the raw data in the Pchatbot dataset can be publicly accessed on the original websites, the raw datasets of Pchatbot are vulnerable to privacy leakage. The reason is that social platforms have ubiquitous private information such as phone numbers, emails, and social media accounts. To protect privacy, these texts are either replaced by indistinguishable marks or deleted depending on whether semantics would be undermined. 

%Besides, data noises are also ubiquitous in the raw datasets. To , we remove meaningless conversations including hashtags, URLs, emoticons, text duplication, and multi-languages using heuristic methods. 

%\dou{we cannot easy understand what kinds of ``challenge'' we have meet when processing the data, i.e., what is the key contribution}

\begin{table*}[t]
\small
\newcommand{\tabincell}[2]{\begin{tabular}{@{}#1@{}}#2\end{tabular}}
\centering 
\caption{Statistics of existing dialogue corpora and Pchatbot. `-' means not being mentioned in corresponding papers. Dialogues means sessions in multi-turn conversations or pairs in single-turn conversations. Utterances means sentences in the dataset. PchatbotW and PchatbotL are the subsets of Pchatbot which we introduce in this paper. }
\begin{tabular}{lrrrrrr}
\toprule
\textbf{Dataset}&\textbf{\# Dialogues} & \textbf{\# Utterances} & \textbf{\# Words} &\textbf{Language} & \textbf{Source} & \textbf{Personalized Info.} \\ \midrule
Twitter Corpus~\cite{Ritter_TwitterV1_2010} & 1,300,000 & 3,000,000 & - & English & Twitter & None \\
% \midrule 
{PERSONA-CHAT~\cite{Zhang_dog_2018}} & 10,981 & 164,356 & - & English & Crowdsourcing & Persona descriptions \\
% \midrule 
{Reddit Corpus~\cite{Samuel_fb_2018}} & 700,000,000 & 1,400,000,000 & - & English & Reddit & User ID \& Timestamp \\
% \midrule
{STC Data~\cite{Wang_STC_2013}}& 38,016 &  618,104 & 15,592,143 & Chinese & Weibo & None\\
% \midrule 
{Noah NRM Data~\cite{Shang_Noah_2015}} & {4,435,959} & 8,871,918 & - & Chinese & Weibo & None\\
% \midrule 
{Douban Conversation Corpus~\cite{Wu_Douban_2017}}  & 1,060,000 & 7,092,000 & 131,747,880 & Chinese & Douban & None \\
% \midrule 
{Personality Assignment Dataset~\cite{Huang_Chat_2018}} & 9,697,651 & 19,395,302 & 166,598,270 & Chinese & Weibo & Key-value pair profile \\
\midrule
{PchatbotW} (Our) &  {139,448,339} & {278,896,678} & {8,512,945,238} & Chinese & Weibo & User ID \& Timestamp \\
% \midrule 
{PchatbotL} (Our) & {59,427,457} & {118,854,914} & {3,031,617,497} & Chinese & Judicial Forums & User ID \& Timestamp \\
%{PchatbotW-1} & {13,992,870} & {27,985,740} & {855,005,996} &
%{\tabincell{c}{ Partition 1 of PchatbotW.}} \\
%\hline
%{PchatbotL-1} & {5,523,160} & {11,046,320} & {284,099,064} &
%{\tabincell{c}{Partition 1 of PchatbotL.}} \\
\bottomrule
\end{tabular}
\label{tab:scale}
\vspace{-10px}
\end{table*}

% \dou{you need to introduce your efforts on processing the data}

%  Intuitively, a larger dataset is expected to bring greater enhancement to dialogue agents, and we hope Pchatbot can bring new opportunities for improving quality of dialogue models. 
% \dou{Pchatbot is not the largest, so we cannot just emphasize the size of the dataset. we need something more than scale. }

Pchatbot is a ready-to-use and well-documented dataset. It is licensed under CC BY-NC\footnote{https://creativecommons.org/licenses/by-nc/2.0/}. Researchers are required to fill in an application form to obtain the dataset files\footnote{https://github.com/qhjqhj00/Pchatbot/blob/main/application.pdf}. For both subsets PchatbotL and PchatbotW, we provide three release versions: (1) the cleaned dataset; (2) the standard dataset for generation-based chatbot; (3) the standard dataset for retrieval-based chatbot. Specifically, the cleaned dataset contains all information of the raw datasets after cleaning processes (\eg, deduplication, anonymization, etc.). 
% For convenient use, the cleaned datasets are evenly divided into ten partitions (see Section \ref{sec:par}). 
The standard datasets are constructed from the cleaned dataset and can be directly used for the corresponding task. For example, we prepare response candidates to each post in the retrieval-based standard dataset. Along with the dataset, we provide tools and tutorials that are used to load, clean, aggregate, and construct the datasets. 

%With the tutorial, user can quickly go through the pipeline 

%With these tools, 

%\dou{can you give a little bit more details for this?}
%\dou{I rembered that we have some data partitions like ClueWeb.}

%\dou{we cannot understand this. you need to mention the difference between them and the specific information contained in them to let reviewers know why they are designed for generation and retrieval. To me, we don't need these two datasets}

\begin{table}[t]
\centering
\small
\caption{Statistics of the standard PchatbotW datasets. }
\begin{tabular}{lrr}
\toprule
& \textbf{PchatbotW-R} & \textbf{PchatbotW-G} \\ 
\midrule
\# Users & 420,000 & 300,000 \\
Avg. history length & 32.3 & 11.4\\
Avg. \# words of post & 24.9 & 22.9\\
Avg. \# words of response & 10.1 & 9.6\\
\# Response candidates & 10 & N/A \\
\# Training samples & 3,000,000 & 2,707,880\\
\# Validation samples & 600,000 & 600,000 \\
\# Testing samples & 600,000 & 600,000\\
\bottomrule
\end{tabular}
\label{tab:dataset}
\vspace{-10px}
\end{table}

% \dou{note that}. 

We experiment with both generation-based and retrieval-based dialogue models on the corresponding standard datasets to provide benchmark that can be used for comparison in the further study. 
% \yutao{We only test one of the application that using the dialogue history as implicit profile. There may be other possible applications, for example, methods using user ID embeddings or general dialogue generation/retrieval.}
Experimental results also verify the advantages of the availability of user IDs and the large-scale data can improve the performance.
%\yutao{Add more observations. For example, implicit profile is effective in training a personalized method and the large-scale data can improve the performance.}
%\dou{we say Applications. Don't make this paper be similar to full papers}

% \dou{please clearly list the advantage of the datasets, and its possible applications}

In summary, the advantages of Pchatbot are as follows:

(1) To the best of our knowledge, the Pchatbot dataset is the largest Chinese dialogue dataset. Neural dialogue models might gain substantial improvements using such a large-scale dataset;

(2) Pchatbot dataset contains two subsets, namely PchatbotW and PchatbotL. The two subsets are dedicated to the open-domain and professional domain (judicial domain), respectively. Such diversity could broaden the application domains of dialogue chatbots.

%We introduce the  The two subsets lay on open domain and judicial domain respectively. The two subsets are significantly larger than existing similar datasets. 

(3) We include anonymized user IDs and timestamps in Pchatbot. This will greatly enlarge the potentiality for developing personalized dialogue agents that learn implicit user profiles from the user's dialogue history.

% . Notably, we have user information for both side of . So you have the possibility of modeling both the current (chatbot) and the dialog target (user). So, do not simply introduce personality

(4) We benchmark several state-of-the-art dialogue models for both generation-based and retrieval-based. Experimental results can be used for comparison in future study. 

% (4) The Pchatbot dataset enlightens a new research pathway for personalized chatbot. Instead of explicit user profile, models can directly learn an implicit user profile from the user's dialogue history. \dou{it is just a possible application, we need to discuss more}

%The Pchatbot dataset will be released upon the acceptance of the paper. We will also release all codes used for data processing and the algorithms implemented in our preliminary experiments.

\section{Related Work}
% With the development of dialogue systems, lots of dialogue datasets have been released. Domain-specific datasets~\cite{DSTC_jason_2013, DBLP:conf/iclr/BordesBW17,Lowe_UbuntuV2_2015} and open-domain datasets~\cite{Ritter_TwitterV1_2010,Sordoni_twitterV2_2015,Wu_Douban_2017,Wang_STC_2013,Cornell_Danescu_2011} are two common types of datasets.

High-quality dialogue systems need large-scale dialogue datasets for training. However, it takes a lot of manual labor and time to collect real human conversation data. In recent years, scholars use post-comments to simulate human dialogue and have published a series of datasets. These datasets can be divided into domain-specific datasets~\cite{DSTC_jason_2013, DBLP:conf/iclr/BordesBW17,Lowe_UbuntuV2_2015} and open-domain datasets~\cite{Ritter_TwitterV1_2010,Sordoni_twitterV2_2015,Wu_Douban_2017,Wang_STC_2013,Cornell_Danescu_2011}. ~\citet{Lowe_UbuntuV2_2015} used Ubuntu Chat Logs to build the Ubuntu dialogue corpus with 930,000 dialogues.
~\citet{JD_data_2018} constructed the JD Customer Service Corpus including 435,005 dialogues based on customer service dialogues from JD.com. These domain-specific datasets can be used to build task-oriented dialogue systems~\cite{DSTC_jason_2013, DBLP:conf/iclr/BordesBW17}.

% Domain-specific datasets are usually used to predict the goal of users in task-oriented dialogue systems~\cite{DSTC_jason_2013, DBLP:conf/iclr/BordesBW17}. Other works collect chat logs from a platform of specific domain~\cite{Lowe_UbuntuV2_2015,JD_data_2018}.
% Since topics of the platforms are in the same domain, the complexity of modeling conversations can be lowered.

% Recently, some researchers create open-domain datasets from movie subtitles~\cite{Cornell_Danescu_2011, Lison_Open_2018,DBLP:conf/acl/ZhuSDNZ20}. However, many conversations are monologues or related to the specific movie scene, which are not suitable for dialogue systems. Other open-domain datasets are constructed from social media networks such as Twitter, Weibo, and Douban~\cite{Ritter_TwitterV1_2010, Sordoni_twitterV2_2015, Wang_STC_2013, Shang_Noah_2015, Wu_Douban_2017}. Nevertheless, with the development of data-driven neural networks, the scale of data is still one of the bottlenecks of dialogue systems.

Open-domain datasets contain conversation data for open topics. Due to the characteristics of social media, the text published by users on social networks is close to real human conversation. In recent years, researchers have constructed some open-domain datasets from social media, such as Twitter~\cite{Ritter_TwitterV1_2010, Sordoni_twitterV2_2015}, Weibo~\cite{Wang_STC_2013, Shang_Noah_2015}, and Douban~\cite{Wu_Douban_2017}. However, we argue that the scopes of these datasets are not enough to train data-driven dialogue systems.

As discussed by~\citet{Personal_turning_VinyalsL15}, it is still difficult for current dialogue systems to pass the Turing test, a major reason is the lack of a coherent personality. In order to train a coherent dialogue system, \citet{Jiwei_Speaker_2016} first attempts to model persona by utilizing user IDs to learn latent variables for representing each user in the Twitter dataset. 
% The Twitter dataset is similar to Pchatbot.
But as far as we know, this dataset has not been made publicly available. To make chatbots maintain a coherent personality, other classic strategies mainly focus on how to endow dialogue systems with a coherent persona by pre-defined attributes or profiles~\cite{Zhang_dog_2018,Samuel_fb_2018,Huang_Chat_2018}.
These works restrict persona in a collection of attributes or texts, which ignore the language behavior and interaction style of a person.

In this work, we construct Pchatbot, a large-scale dataset with personalized information, to solve the mentioned issues. Pchatbot has two subsets from open domain and specific domain respectively.
%, named PchatbotW and PchatbotL. PchatbotW has 130 million pairs collected from open-domain Weibo data, while PchatbotL has 59 million pairs collected from judicial professional specific-domain.

Table~\ref{tab:scale} shows the data scales of Pchatbot and other datasets. In addition, all posts and responses of Pchatbot are attached with user IDs and timestamps, which can be used to learn not only persona profiles but interaction styles from the users' dialogue histories.

\section{Pchatbot Dialogue Dataset}
Pchatbot dataset is sourced from public websites. It has two subsets from Weibo and judicial forums respectively. Raw data are normalized by removing invalid texts and duplications. Privacy information in raw data is also anonymized using  indistinguishable placeholders.
\subsection{Dataset Construction}
Each item of raw data in the dataset is started by a post made by one user and multiple responses then follow. We extract the post-response pairs from the original threads.

% \dou{you need to at least mention you extract post-response pairs from the original threads...}

\subsubsection{General Preprocessing Pipeline}
Since the Pchatbot dataset is collected from social media and forums, private information such as homepage, telephone, email, ID card number, and social media account, is ubiquitous. Besides, there are also many sensitive words such as pornography, abuse, and political words.
Therefore, we design a preprocessing pipeline to deal with the raw data, which contains the following four steps:

(1) \textbf{Anonymization.} We replace private information in the data with placeholders using either rule-based methods or information extraction models. Specifically, we use regex expressions to recognize texts such as email, phone numbers, and account numbers. We use NER models to extract entities like names and addresses.

% \yutao{is there any reference for this? [Following xxx]}

(2) \textbf{Filtering Sensitive Words.}  The sensitive words are detected by the matching method with a refined sensitive word list\footnote{\url{https://github.com/fighting41love/funNLP}\label{fun}}. As sensitive words are also very important in terms of semantics, simply replacing them with placeholders will undermine the completeness of the sentences. Therefore, we directly filter out all (post, response) pairs with sensitive words.
% if  are detected,  would be filtered.

(3) \textbf{Filtering Utterances by Length.} We clean the utterance whose length is less than 5 or more than 200 because the short utterances tend to contain limited information, while the long utterances usually have noise. 

(4) \textbf{Word Segmentation.} For Chinese word segmentation, we use Jieba toolkit\footnote{\url{https://github.com/fxsjy/jieba}}. Since Jieba is implemented for general Chinese word segmentation, we introduce a law terminology list\textsuperscript{\ref {fun}} as an extra dictionary for enhancement in PchatbotL.

As the two subsets of Pchatbot are from different sources, in addition to the general preprocessing pipeline, the detailed preprocessing strategies and descriptions are introduced as follows.

\subsubsection{PchatbotW}
In China, Weibo is one of the most popular social media platforms for users to discuss various topics and express their opinions. The basic function of this platform is very similar to Twitter. We crawl the publicly available Weibo posts and their comments across one year (from September 10, 2018 to September 10, 2019). Then, we randomly select about 23M users, and get their conversation histories. As a result, we obtain 341 million (post, response) pairs in total.

Due to the nature of social media, posts and responses published by Weibo users are extremely diverse that cover various aspects of daily life. Therefore, interactions between users can be considered as daily casual conversations in the open domain. Since Weibo text is in a casual manner containing a lot of noise, to improve the quality of the data, we do the following data cleaning operations in addition to the general preprocessing pipeline:

(1) \textbf{Removing Hashtags.} Users like to tag their contents with related topics by hashtags, which usually consist of several independent words or summaries wrapped by `\#'. Since splicing hashtag text into contents will affect the semantic coherence, we remove the hashtags from the content.
(2) \textbf{Removing URLs.} Users' posts and responses sometimes contain multimedia content, images, videos, and other web pages. They are converted to URLs in Weibo. We also remove these URLs from the content.
(3) \textbf{Removing Emoticons.} On the Weibo platform, users can use emoticons (including emoji and kaomoji) to convey their emotions. Emoticons consist of various symbols, which introduce noises to dialogues. Therefore, we remove these emoticons by regex and dictionary.
(4) \textbf{Handling Mentions.} On the platform, users can also use `@nickname' to mention or notice other users. When users comment or repost others, `Reply @nickname:' and `//@nickname:' will be automatically added into the users' contents. These mentions serve as reminders and often have little relevance to the users' content. So, we remove them to guarantee the consistency of utterances.
(5) \textbf{Handling Duplicate Texts.} Duplicate texts appear in different granularities. For word-level duplication, the duplicated Chinese characters are normalized to two characters. For example, ``\begin{CJK*}{UTF8}{gkai} 太好笑了，哈哈哈哈哈\end{CJK*}'' (``That's so funny. hahahahaha'') will be normalized as ``\begin{CJK*}{UTF8}{gkai}太好笑了，哈哈\end{CJK*}'' (``That's so funny. haha''). As for response-level duplication under a post, they are usually caused by different users sending the same responses. Duplicate responses under a post reduce the varieties of interactions, so we remove those occurring more than three times. Furthermore, we also find utterance-level duplication in the dataset. Duplicated utterances in the entire dataset may affect the balance of the dataset. Models are reported to generate general responses when being trained on a large number of duplicate utterances~\cite{Jiwei_Speaker_2016}. Specifically, in PchatbotW, we limit the frequency of the same utterances as 10,000. For utterances that have a frequency over 10,000, we randomly remove the over part.
% \dou{what we will do for other utterances not within 10,000?}
(6) \textbf{Multi-languages.} Due to the diversity of Weibo, some users' content contains multiple languages. We remove samples containing more than 70\% content that is in other languages.

% \yutao{add purpose for constructing standard datasets.}
For fair comparison in future work, we construct two standard datasets from PchatbotW to evaluate dialogue models, namely PchatbotW-R and PchatbotW-G, which are used for retrieval-based and generation-based tasks, respectively. Concretely, following previous works that construct retrieval-based datasets, in PchatbotW-R, we retrieve 10 response candidates for each data sample. The generation-based dataset PchatbotW-G is directly derived from the PchatbotW. Statistics for the two datasets are shown in Table \ref{tab:dataset}.

% \dou{why just using the first partition? NO!}

% Example utterances containing texts that needed to be cleaned are shown in Table~\ref{tab:examples_weibo}. Regex patterns used in the data cleaning step are shown in Table~\ref{tab:reglist}. 

\begin{figure}[t!]
    \centering
    \includegraphics[width=\linewidth]{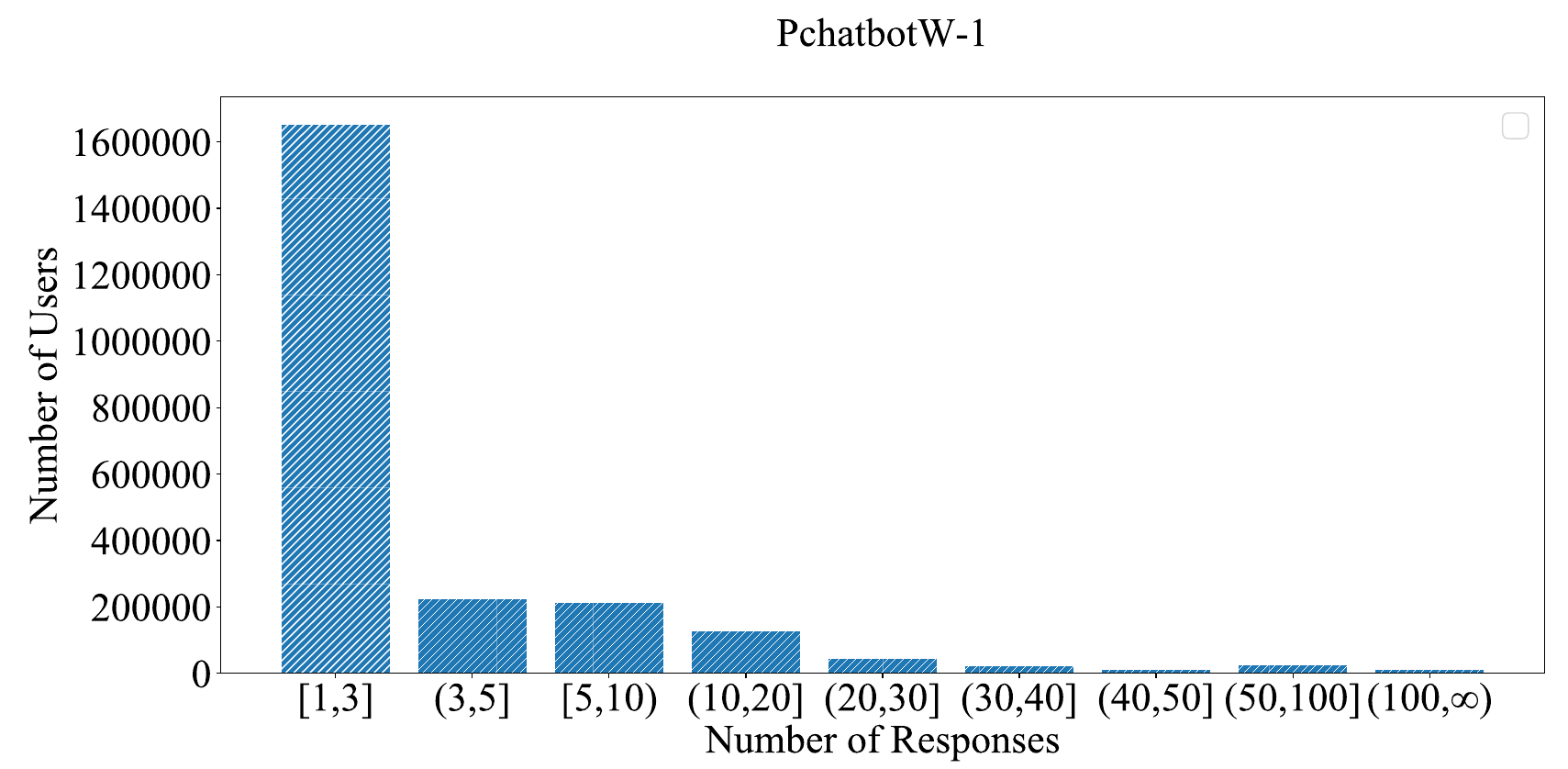}
    \caption{Distribution of users' number with different scopes of responses on PchatbotW-1}
    \label{fig:W1}
    \vspace{-10px}
\end{figure}
\begin{figure}[t!]
    \centering
    \includegraphics[width=\linewidth]{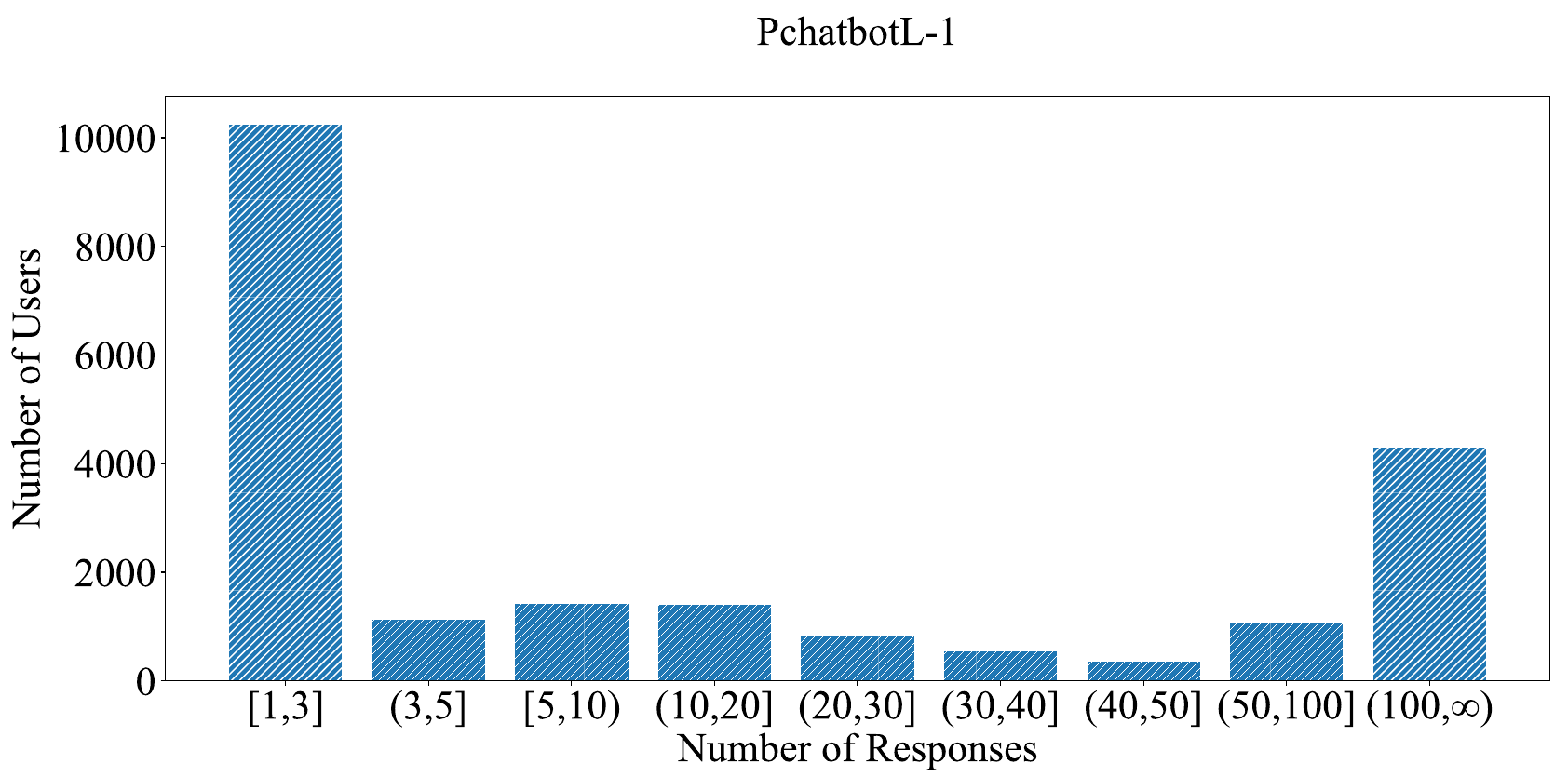}
    \caption{Distribution of users' number with different scopes of responses on PchatbotL-1}
    \label{fig:L1}
    \vspace{-10px}
\end{figure}

\begin{CJK*}{UTF8}{gkai}
\begin{table}[!t]
\centering
\small
\caption{An example of data in Pchatbot}
\setlength{\tabcolsep}{0.7mm}{
\begin{tabular}{ll}
\toprule
\textbf{Field}   & \textbf{Content}   \\
\midrule
Post  & 下冰雹了!真刺激!\;(Hailing! It's really exciting!) \\
% \hline
Post user ID  & 5821954   \\
% \hline
Post timestamp & 634760927 \\
% \midrule
Response   & 出去感受更刺激\;(It's more exciting to go out.) \\
% \hline
Response user ID   & 592445\\
% \hline
Response timestamp & 634812525 \\
% \hline
Partition index (1-10) & 1 \\
% \hline
Train/Dev/Test (0/1/2)   & 0 \\ 
\bottomrule
\end{tabular}
}
\label{tab:data_scheme}
\vspace{-10px}
\end{table}
\end{CJK*}

\subsubsection{PchatbotL}
Judicial forums are professional platforms that open to users for consultation and discussion in the judicial domain. People can seek legal aid from lawyers or solve the legal problems of other users in the judicial forums.

We crawl around 59 million post-response pairs from 5 judicial websites, including 66law.cn, findlaw.cn, lawtime.cn, 110.com, and 9ask.cn, from October 2003 to February 2017. Since the data of the judicial forums are questions from users and answers from lawyers, topics mainly focus on the legal domain. Post-response pairs are almost of high quality so that only basic preprocesses are needed. More information for the dataset can be found on the release page. 
 
\begin{table*}[t]
\centering
%\small
\caption{Experimental results on the PchatbotW-R and PchatbotW-G.}
\begin{tabular}{lllll|llllll} 
\toprule
% \multirow{2}*{\textbf{Type}} & 
%   & \multicolumn{4}{c}{PchatbotW-R} & &\multicolumn{5}{c}{PchatbotW-G} \\
%   \cmidrule(lr){2-5} \cmidrule(lr){7-11}
  PchatbotW-R & $\mathbf{R_{10}@1}$ &  $\mathbf{R_{10}@2}$ & $\mathbf{R_{10}@5}$ & \textbf{MRR}  & PchatbotW-G &\textbf{BLEU-1} & \textbf{ROUGE-L} & \textbf{Dist-1} & \textbf{Dist-2} & \textbf{P-F1}\\  
  \midrule
  (1) Conv-KNRM   & 0.323 & 0.520 & 0.893 & 0.538 & (2) Seq2Seq &4.889&7.594&0.299&3.404&0.771   \\ 
  (1) DAM &0.438 & 0.644 & 0.966 & 0.635 &  (2) Speaker &3.958&5.580&0.951&29.780&1.534  \\ 
  (1) IOI & \textbf{0.442} & \textbf{0.651} & \textbf{0.969}  & \textbf{0.639} &  (2) PersonaWAE &1.945&\textbf{9.064}&0.523&8.549&\textbf{6.408}  \\
  (1) RSM-DCK  & 0.428 & 0.627 & 0.947 & 0.623 &  (2) DialoGPT &\textbf{5.038}&7.458&\textbf{13.995}&\textbf{52.674}&3.562 \\

\bottomrule
\end{tabular}
\label{tab:res}
%\vspace{-10px}
\end{table*}

\subsection{Data Partition}
\label{sec:par}

The original Pchatbot dataset is very large. For convenient use, we divide Pchatbot into 10 partitions evenly according to the user IDs in responses. Each partition has a similar size of (post, response) pairs. PchatbotL-1 and PchatbotW-1 are the first partitions of PchatbotL and PchatbotW, respectively. Figure~\ref{fig:W1} and Figure~\ref{fig:L1} show the distribution of the length of users' dialogue history of PchatbotW-1 and PchatbotL-1, respectively. We find that the users in the PchatbotL tend to have more responses than users in the PchatbotW. The reason is that repliers in PchatbotL are usually professional lawyers who routinely provide judicial help in the forums. Other partitions have similar distributions. We also divide the datasets into train/dev/test sets. In the division of train/dev/test set, given a user, we ensure that the time of its records in the dev-set and test-set are behind the records in the train-set by using timestamps.

%And we limit the number of dev and test sets below 3 and 15 for PchatbotL and PchatbotW separately in a result of their differences in users' average responses\dou{this sentence is hard to understand. And why?} .

% \yutao{I think we only need to split the data for the standard dataset? By the way, we also need to construct a standard legal dataset in the future.}
 
\subsection{Data Format and Statistics}
\begin{table}[t]
\small
\centering
\caption{Detailed statistics. PchatbotW-1 and PchatbotL-1 are the 10\% partitions of the corresponding subsets.}
\setlength{\tabcolsep}{0.5mm}{
\begin{tabular}{lrrrr}
\toprule
 & \textbf{PchatbotW} & \textbf{PchatbotL} & \textbf{PchatbotW-1}& \textbf{PchatbotL-1} \\ \midrule
{\# Posts} &5,319,596 & 20,145,956 &3,597,407 &4,662,911\\ 
{\# Responses} &139,448,339& 59,427,457 & 13,992,870& 5,523,160\\ 
% \midrule
{\# Users in posts}  &772,002 &5,203,345& 417,294& 1,107,989\\
{\# Users in resp.}  &23,408,367 & 203,636 &2,340,837 &20,364\\ 
% \midrule
{Avg. \# resp. / post}&26.214 &2.950 &3.890 &1.184\\ 
{Max. \# resp. / post}&525 &120 & 136 & 26\\ 
% \midrule
{Vocabulary Size}  & 9,148,532&1,329,930 & 3,447,433& 551,071\\ 
{Avg. \# words of post} & 49.37 &36.88 & 49.40& 37.26\\ 
{Avg. \# words of resp.} & 11.68 &14.08 &11.70 &14.13 \\ \bottomrule
\end{tabular}
}
\label{tab:stat_Pchatbot}
\vspace{-10px}
\end{table}

Pchatbot's schema is shown in Table~\ref{tab:data_scheme}. Each record of Pchatbot includes 8 fields: post, post user ID, post timestamp, response, response user ID, response timestamp, partition, and train/dev/test identity. User IDs and timestamps are attached in the Pchatbot dataset for each post or response. User IDs can be used to distinguish the publisher of each post or response. Timestamps provide temporal information that can be used to build a historical response sequence for each user. The historical sequence could help to train dialogue models that imitate the speaking style of specific users. 

In Table~\ref{tab:stat_Pchatbot}, we show the statistics of the Pchatbot dataset.
We find that the number of users who comment (23,408,367) is significantly larger than those who post (772,002) in PchatbotW. However, in PchatbotL, the number of users who comment (203,636) is much smaller than the number of users who post (5,203,345). We attribute this to the differences between the two platforms. Social media users are more willing to engage in interactions, while users from judicial forums tend to ask legal questions. Besides, the number of lawyers who answer legal questions in judicial forums is limited. 

The scale of the Pchatbot dataset significantly outperforms previous Chinese datasets for dialogue generation. To be concrete, PchatbotW contains 5,319,596 posts and more than 139 million (post, response) pairs. PchatbotL contains 20,145,956 posts and more than 59 million (post, response) pairs. The largest dataset before has only less than 10 million (post, response) pairs. With such scales, performance improvement for data-driven neural dialogue models can be almost guaranteed. Pchatbot dataset provides sufficient valid responses as ground-truth for a post. On average, each post has 26 responses in PchatbotW. This helps to establish dialogue assessment indicators at the discourse level. 

%\dou{within the paper, you never mentioned Reddit?? You can claim our dataset is in Chinese....}
%

\section{Application and Analysis}
The Pchatbot dataset can be used in a wide range of applications of the dialogue system.  In this section, we conduct preliminary studies on the effectiveness of the dataset. We first benchmark state-of-the-art models over Pchatbot for comparison in future work.  We also investigate the effectiveness of the scale of training data and the length of dialogue history, respectively. 
% \yutao{recall that our experiment is only for personalized dialogue}

\subsection{Settings and Evaluation Metrics}
In our benchmark experiments, we keep the parameter setting the same as described in the corresponding papers except that we replace the word embeddings with ours. We will release codes for benchmark models on the release page.
We also provide pre-trained language models including GloVe~\cite{GloVe_PenningtonSM14}, BPE~\cite{BPE_SennrichHB16a}, Fasttext~\cite{Fasttext} which are trained on the dataset. These pre-trained models can be downloaded on the release page.

%we use 4 layers GRU models with Adam optimizer. Hidden size for each layer is set to 1,024. Batch size is set to 128, embedding size is set to 100. Learning rate is set to 0.0001 and decay factor is 0.995. Parameters are initialized by scope [-0.8,0.8]. Gradients clip threshold is set to 5. Vocabulary size is set to 40,000. Dropout rate is set to 0.3. Beam width is set to 10 for beam search.

For retrieval-based dialogue models, we use $\mathbf{R_n@k}$ (recall at position $k$ in $n$ candidates) and \textbf{MRR} (Mean Reciprocal Rank) to measure the model's ability to select a personalized response from all candidates. For generation-based dialogue models, we use the \textbf{BLEU}~\cite{Salim_BLEU_2002} metric which is widely used to evaluate the model-generated responses.  Besides, we use \textbf{Distinct-1/2}  proposed in~\cite{Jiwei_distinct_2016} to evaluate the diversity of responses generated by the model. To measure the personality consistency of generation-based models, we use \textbf{P-F1} as an evaluation metric~\cite{lian2019learning}.

%In a conversation, the same post can have a variety of replies, so the automatic metrics have great limitations in evaluating dialogue~\cite{Eval_2016}. Therefore, for each dataset, we sample 300 responses generated by each model, and manually label them according to fluency, correlation and personality. The scoring criteria is as follows: 0(not fluently), 1(fluently but irrelevant), 2(relevant but generic), 3(fit for post), and 4(like a person).

\subsection{Benchmark Models}
We experiment with the following models\footnote{We will continue updating other benchmark results on the project page.}:

(1) Retrieval-based models: \textbf{Conv-KNRM}~\cite{10.1145/3159652.3159659}: Single-turn dialogue model that uses CNN to capture n-gram features; \textbf{DAM}~\cite{zhou-etal-2018-multi}: Multi-turn dialogue model that takes the user's dialogue history as context to construct multi-level text segment representations with stacked self-attention; \textbf{IOI}~\cite{tao-etal-2019-one}: Multi-turn dialogue model that captures deep interactive matching features between response and utterance in the conversation context; \textbf{RSM-DCK}~\cite{10.1145/3340531.3411967}: Knowledge enhanced multi-turn dialogue model that considers persona descriptions as external knowledge.

%The model pre-select document and context using recent context as key. It then conduct matching between response-context and response-document. It finally uses Bi-LSTM to aggregate matching features.

(2) Generation-based models: \textbf{Seq2Seq}~\cite{Ilya_S2S_2014, Luong_attention_2015}: Single-turn dialogue model that uses RNN-based encoder-decoder to generate responses; \textbf{Speaker Model}~\cite{Jiwei_Speaker_2016}: Single-turn personalized dialogue model that utilizes user IDs to learn user embeddings; \textbf{PersonaWAE}~\cite{yanrui_pwae_2019}: Single-turn personalized dialogue model that samples the personalized vector from a personalized Gaussian mixture distribution and uses it to guide the response generation; \textbf{DialoGPT}~\cite{zhang2020dialogpt}: Large-scale dialogue model that concatenates the utterances as a long sequence and learns to generate the response.

\begin{figure}[t]
    \centering
    \includegraphics[width=0.45\textwidth]{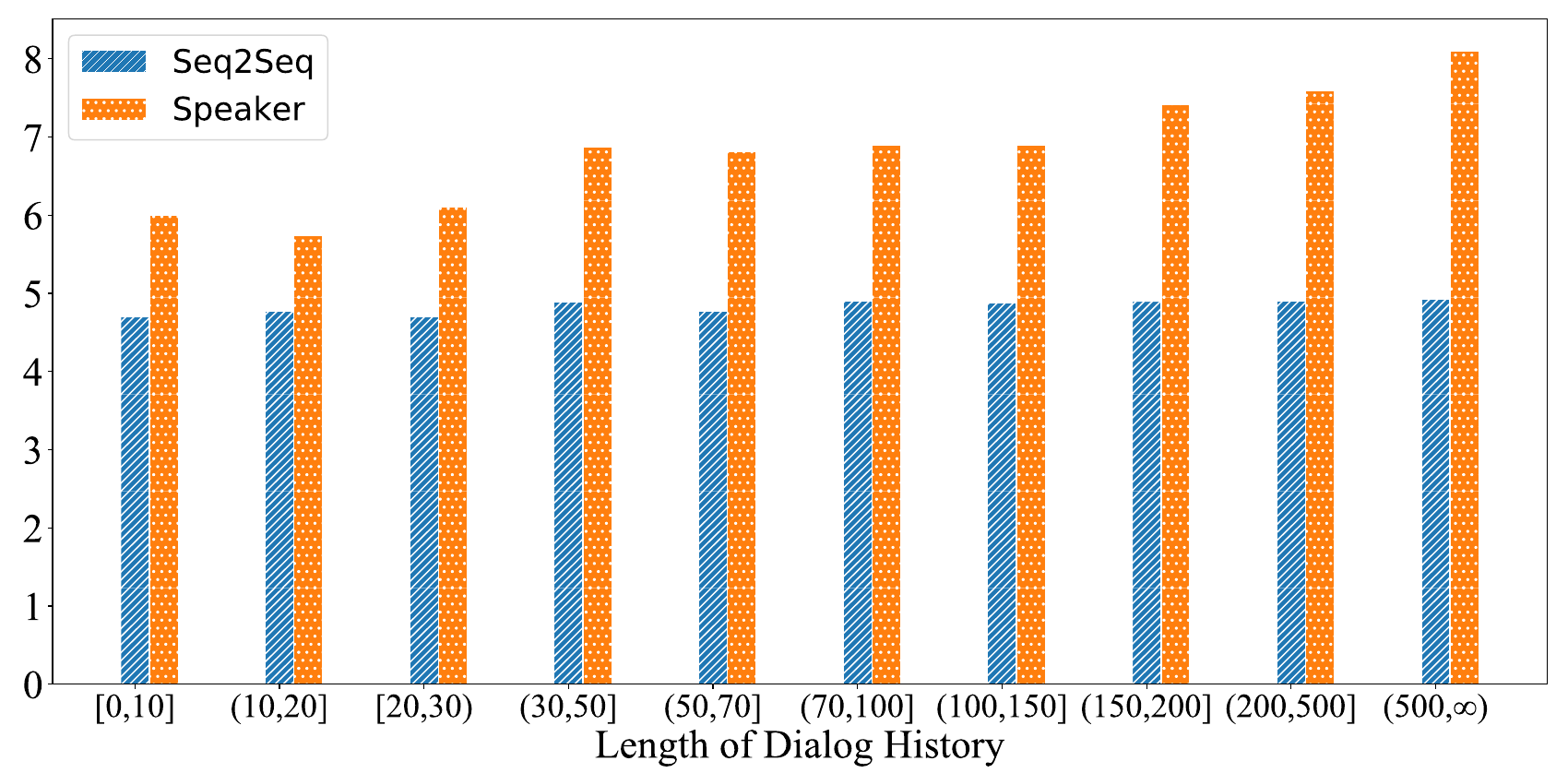}
    \caption{Effect of the length of dialogue history.}
    \label{fig:bleu1}
    \vspace{-10px}
\end{figure}

%\begin{figure}[t]
%    \centering
%    \includegraphics[width=0.35\textwidth]{./figure/length.pdf}
%    \caption{Comparison of BLEU-1 on two models on PchatbotL.}
%    \label{fig:bleu1}
%\end{figure}

\subsection{Benchmark Experiments}

%In this section, we first evaluate the performances of state-of-the-art models on the PchatbotW-R and PchatbotW-G datasets. Then, we investigate the impact of data scale and the length of the user's dialogue history, respectively.

\subsubsection{Results} 

Table \ref{tab:res} shows the experimental results. For the retrieval-based models, we have the following findings: (1) single-turn model (Conv-KNRM) performs worse than others. The potential reason is the lack of conversational context which refers to the user's dialogue history in our scenario; (2) multi-turn model (DAM and IOI) performs better than others illustrating that using user's dialogue history as context can effectively booster the performance of ad-hoc matching; (3) for the knowledge-enhanced model (RSM-DCK), following \citet{zhang2020dialogpt}, we apply heuristic methods to generate external knowledge. But the effectiveness of such external knowledge is limited, which enlightens the need to directly model the user's personality from the user's dialogue history.

Generation-based models lead to similar findings: (1) the personalized models (Speaker and PersonaWAE) significantly outperforms the Seq2Seq model regarding Dist-1, Dist-2, and P-F1, which demonstrates that modeling user indeed leads to generating informative and personalized responses; (2) large-scale model (DialoGPT) obtains the best results for its great generalization ability brought by large-scale parameters. However, it performs worse than PersonaWAE regarding P-F1. 

In general, using the user's dialogue history as context indeed improves dialogue models' performances especially for personalized metrics. Besides, personalized models that use either explicit user profiles or user embeddings fail to fully explore the personalized information that is hidden under the user's dialogue history. Pchatbot dataset enables designing models that can directly learn implicit user profiles. Such models are more useful regarding personalized chatbots, especially in a practical scenario.

%the personalized model (PersonaWAE) achieves the best performance regarding ROUGE-L, which can be explained by the use of BOW Loss. Besides, PersonaWAE achieves the best P-F1, which further illustrates the benefits of user modelling;

\subsubsection{Impact of the Length of Dialogue History}

We evaluate the quality of responses generated for users with different lengths of dialogue history. For personalized chatbots, we expect the chatbots generate responses that are close to the original responses. Thus, we choose BLEU as the key indicator. We conduct experiments on a personalized dialogue model (Speaker) and a non-personalized chatbot model (Seq2Seq). The results are shown in Figure~\ref{fig:bleu1}. From the figure we can find that: (1) with more dialogue history, the two models' performances continue to increase; (2) the discrepancy of BLEU scores between the two models gradually increases. In other words, personalized dialogue models can benefit more than non-personalized models.

%This further confirms the usefulness of historical user conversations for response generation. With more historical conversations, the model could learn more stable user persona, and tends to generate meaningful and consistent responses closer to the original responses. 

\subsubsection{Impact of the Scale of Data}\label{sec:expr_of_scale}
We evaluate the effectiveness of dataset scale by conducting experiments on five subsets of different sizes using the Seq2Seq model. We construct these subsets by merging the partitions. Specifically, we use partition-1 as the smallest dataset and add partition-2 to partition-5 successively to construct bigger datasets.

\begin{figure}[t]
    \centering
    \includegraphics[width=0.45\textwidth]{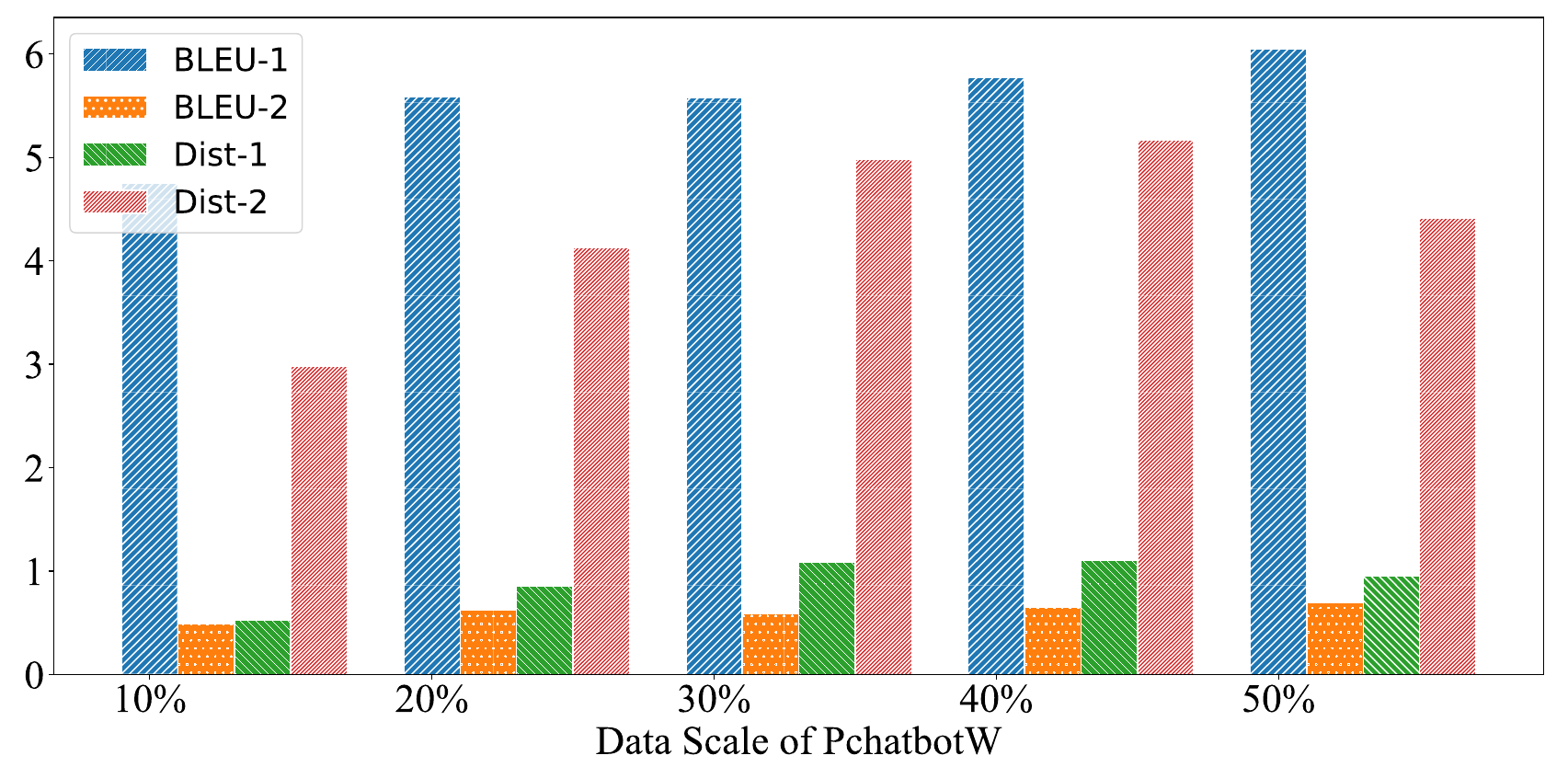}
    \caption{Effect of data scale.}
    \label{fig:ratio}
    \vspace{-10px}
\end{figure}

Experimental results of incremental scale are shown in Figure \ref{fig:ratio}. The results show that with the increasing of the training data size, the model's performance has a growing trend across all metrics. It confirms that using more training data helps to improve the model's effectiveness. Besides, we find that the diversity metrics(Distinct-K) turn goes down when using 50\% of data. We attribute the phenomenon to that with more training data, the Seq2Seq model prefers to generate similar and generic responses~\cite{Jiwei_Speaker_2016}.

%However, we also find that the diversity metrics(Distinct-K) in different subsets have no obvious discrepancy. We attribute this to that our smallest dataset also has a relatively big scale, which highlights the common disadvantage in generation-based models: preferring to generate similar and generic responses~\cite{Jiwei_Speaker_2016}.

\section{Conclusion and Future Work}
In this paper, we introduce the Pchatbot dataset that has two subsets from the open domain and judicial domain respectively, namely PchatbotW and PchatbotL. All posts and responses in Pchatbot are attached with anonymous user IDs as well as timestamps, which greatly broadens the potentialities of a personalized chatbot. Besides, the scale of the Pchatbot dataset is significantly larger than previous datasets and this further enhances the capacity of intelligent dialogue agents. We evaluate the Pchatbot dataset with several baseline models and experimental results demonstrate the great advantages triggered by user IDs and large scale. The Pchatbot dataset and corresponding codes can be publicly viewed at Github.

\section*{Acknowledgements}
Zhicheng Dou is the corresponding author. This work was supported by National Natural Science Foundation of China No. 61872370 and No. 61832017, Beijing Outstanding Young Scientist Program NO. BJJWZYJH012019100020098, and Shandong Provincial Natural Science Foundation under Grant ZR2019ZD06.

%Personalized chatbot is an interesting research problem. In this paper, we did some preliminary studies on personalized conversation generation on our proposed Pchatbot dataset. Advanced personalized chatbot models are beyond the scope of this paper, and we will explore them in future work.

\clearpage
\bibliographystyle{ACM-Reference-Format}
\bibliography{sample-base}

%%
%% If your work has an appendix, this is the place to put it.

\end{document}